\pdfoutput=1

\documentclass[11pt]{article}

\usepackage[final]{acl}

\usepackage{times}
\usepackage{latexsym}
\usepackage{makecell}

\usepackage[T1]{fontenc}

\usepackage[utf8]{inputenc}

\usepackage{microtype}

\usepackage{inconsolata}

\usepackage{url}            
\usepackage{booktabs}       
\usepackage{amsfonts}       
\usepackage{nicefrac}       
\usepackage{xcolor}         

\definecolor{lightgrayrow}{gray}{0.94}
\definecolor{headgray}{gray}{0.88}
\definecolor{highlightrow}{gray}{0.90}

\usepackage{url}
\usepackage{tabularray}
\UseTblrLibrary{booktabs}
\usepackage{graphicx}
\usepackage{tcolorbox}

\usepackage{booktabs}
\usepackage{multirow}
\usepackage{bm}          

\usepackage{amsmath}
\usepackage{multirow}
\usepackage{colortbl}
\usepackage{booktabs}
\renewcommand\arraystretch{0.95}
\usepackage{wrapfig}
\usepackage{xspace}
\usepackage{arydshln}
\usepackage{subcaption}
\usepackage{paralist}
\usepackage{enumitem}
\usepackage{hyperref}

\usepackage{tikz}
\definecolor{forestgrhl}{RGB}{84, 186, 111}
\definecolor{darkgrhl}{RGB}{136, 219, 158}
\definecolor{greenhl}{RGB}{201, 242, 212}
\definecolor{faintgrhl}{RGB}{244, 255, 241}
\definecolor{darkredhl}{RGB}{232, 173, 161}
\definecolor{redhl}{RGB}{242, 208, 201}
\definecolor{faintredhl}{RGB}{255, 244, 241}
\definecolor{whitehl}{RGB}{255, 255, 255}
\definecolor{yellowhl}{RGB}{255, 217, 164}
\definecolor{forestgreen}{rgb}{0.13, 0.55, 0.13}
\definecolor{bluehl}{RGB}{201, 225, 242}

\title{GraphMed-LT: Patient-Specific Graph Memory with Latent Clinical Thought Refinement for Multi-Turn Medical Conversations}

\author{
  Zhaohan Meng$^{1,2}$,
  Zaiqiao Meng$^{1,3}$,
  Siwei Liu$^{4}$,
  Hao Xu$^{2}$,
  Ke Yuan$^{5,6}$,
  Iadh Ounis$^{1}$   
  \vspace{2mm}\\
  $^{1}$School of Computing Science, University of Glasgow\\
  $^{2}$Brigham and Women's Hospital, Harvard Medical School, Harvard University\\
  $^{3}$Language Technology Lab, University of Cambridge\\
  $^{4}$School of Natural and Computing Science, University of Aberdeen\\
  $^{5}$School of Cancer Sciences, University of Glasgow\\
  $^{6}$Cancer Research UK Scotland Institute\\
  \texttt{z.meng.3@research.gla.ac.uk}\\
}

\begin{document}

\maketitle

\begin{abstract}
Multi-turn medical question answering (QA) aims to model realistic clinical diagnosis, where a doctor gathers patient information across multiple turns of conversation. Existing multi-turn medical conversation systems have shown promising progress, but they often rely on accumulated conversation histories as memory, leaving clinical evidence fragmented across turns. We propose \textbf{GraphMed-LT}, a patient-specific graph memory approach with latent clinical thought refinement for multi-turn medical conversations. GraphMed-LT extracts patient-specific clinical triplets from patient responses, retrieves relevant knowledge triplets, and organises them into an incrementally updated graph memory. The graph memory is projected into graph-conditioned evidence tokens and refined inside a trainable doctor agent through hidden-state feedback, enabling the agent to update its internal clinical context before asking follow-up questions or producing the final answer. Experiments on three multi-turn medical QA benchmarks show that GraphMed-LT consistently outperforms existing multi-turn medical conversation baselines across multiple LLM backbones, achieving up to a 6.3 percentage-point absolute improvement over the strongest baseline. Further analyses show that GraphMed-LT asks more answerable follow-up questions and provides consistent gains across medical specialties.\footnote{The source code for GraphMed-LT is publicly available at \href{https://github.com/ZhaohanM/GraphMed-LT}{https://github.com/ZhaohanM/GraphMed-LT}.}
\end{abstract}

\section{Introduction}

\begin{figure*}[t]
    \centering
    \includegraphics[width=\linewidth]{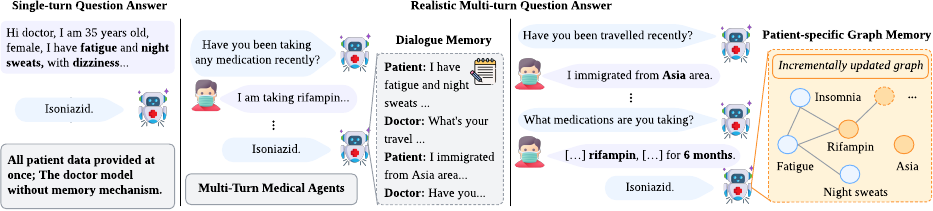}
    \caption{
    \textbf{Single-turn QA vs. Multi-turn QA.}
    \textbf{Left:} In single-turn medical QA, all patient information is provided at once without asking follow-up questions.
    \textbf{Middle:} In multi-turn medical QA, patient information is collected progressively and stored as dialogue history.
    \textbf{Right:} GraphMed-LT organises the acquired patient information into a patient-specific graph memory with latent clinical thought refinement for multi-turn reasoning.
    }
    \label{fig:task}
\end{figure*}

Traditional medical question answering (QA) is typically formulated as a single-turn task, where all patient information is provided at once~\cite{jin2019pubmedqa,pal2022medmcqa}. As illustrated in Fig.~\ref{fig:task}, this differs from multi-turn medical conversations, where patient information is collected progressively through doctor--patient interaction. Although existing approaches have achieved strong performance in single-turn QA, they do not fully reflect real-world clinical practice, where diagnoses are formed through iterative questioning and progressive evidence accumulation~\cite{hendrycks2021measuring}. A central challenge in multi-turn medical conversations is that clinically relevant evidence is distributed across turns, requiring effective integration for reliable multi-hop reasoning~\cite{jin2024large,wang2023can}. This motivates structured mechanisms for organising and integrating clinical evidence acquired across turns.

Existing methods for multi-turn medical reasoning have achieved promising performance by optimising information-seeking behaviour, enabling doctor agents to ask follow-up questions and acquire missing patient evidence before making a decision~\cite{hu2024uncertainty,chen2025cod,meng2026triplet}. MEDIQ~\cite{li2024mediq} established a doctor--patient conversation setting in which a doctor agent actively queries a patient simulator and reasons over accumulated conversation logs. Building on this interactive setting, ProMed~\cite{ding2025promed} further encouraged proactive questioning through reinforcement learning with Shapley Information Gain~\cite{winter2002shapley} rewards. MEDDxAgent~\cite{rose2025meddxagent} introduced a modular differential-diagnosis agent that combines history taking and knowledge retrieval for interactive clinical reasoning. These methods demonstrate the value of active information seeking in multi-turn medical conversations. However, patient evidence is still largely maintained as dialogue history rather than organised into a structured cross-turn memory. Therefore, clinical facts acquired in separate turns remain fragmented, limiting the multi-turn dependencies required for effective multi-hop reasoning.

To address this challenge, we propose GraphMed-LT, a patient-specific graph memory approach with latent clinical thought refinement for multi-turn medical conversations. GraphMed-LT extracts patient-specific triplets from patient responses, retrieves relevant knowledge triplets, and organises them into an incrementally updated graph memory. Knowledge triplets provide a suitable representation for complex biomedical knowledge by decomposing clinical evidence into explicit entity--relation--entity structures, making the underlying reasoning chains easier to retrieve, trace, and verify than in free-form text. By accumulating these triplets in a patient-specific graph memory, GraphMed-LT connects evidence acquired across turns rather than treating it as a dialogue history. The graph memory is projected into graph-conditioned evidence tokens, which are in turn injected into a trainable doctor large language model (LLM) and refined through continuous-space hidden-state feedback before follow-up questioning or final answer prediction. 

To assess whether this design improves both final diagnosis and the intermediate information-seeking process, we evaluate GraphMed-LT on three multi-turn medical conversation datasets: iCRAFT-MD, iMedQA, and iMedMCQA~\cite{johri2023testing,johri2024craft,li2024mediq,pal2022medmcqa}. Experimental results show that GraphMed-LT consistently outperforms seven strong baselines across multiple LLM backbones. Further analyses show that GraphMed-LT asks more answerable follow-up questions and provides consistent gains across medical specialties. These findings indicate that combining patient-specific graph memory with latent clinical thought refinement enables more reliable multi-hop reasoning in multi-turn medical conversations. Our contributions can be summarised as follows:

\begin{itemize}
    \item We propose GraphMed-LT, a patient-specific graph memory approach that incrementally organises extracted patient triplets and retrieved knowledge triplets for multi-turn clinical evidence modelling.
    \item We introduce latent clinical thought refinement, a continuous-space reasoning mechanism that refines graph-conditioned evidence tokens inside a trainable doctor agent through hidden-state feedback.
    \item We conduct extensive experiments across three multi-turn medical conversation datasets, showing that GraphMed-LT consistently outperforms strong baselines across multiple LLM backbones.
\end{itemize}


\begin{figure*}[t]
    \centering
    \includegraphics[width=0.9\linewidth]{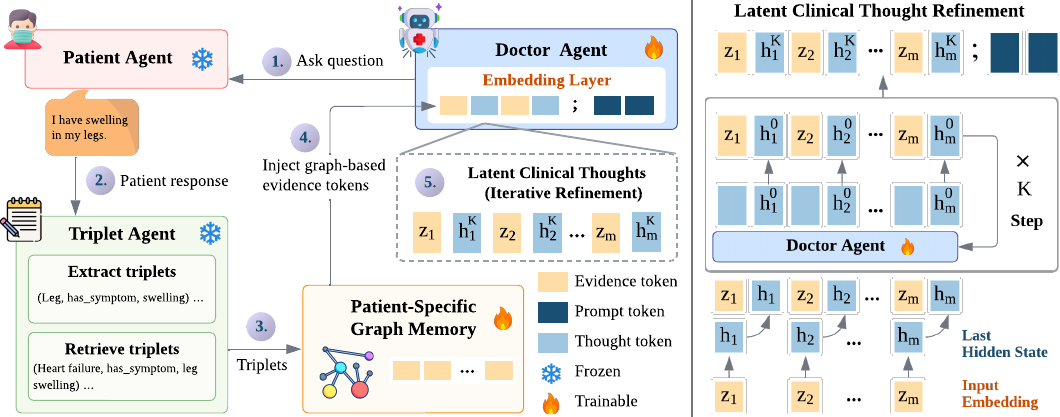}
    \caption{
    \textbf{Left:} The overview of GraphMed-LT follows steps 1--5: the doctor asks a follow-up question, the patient responds, the triplet agent extracts and retrieves triplets, and the patient-specific graph memory projects them into graph-based evidence tokens for the doctor agent.
    \textbf{Right:}
    (1) graph-based evidence tokens $z_i$ are paired with thought tokens $h_i$;
    (2) during the $K$-step loop, $z_i$ remains fixed while $h_i$ is updated through hidden-state feedback;
    (3) the final sequence $[z_1,h_1^K,\ldots,z_m,h_m^K]$ is concatenated with prompt for follow-up questioning or answer prediction.
    }
    \label{fig:main_figure}
\end{figure*}

\section{Methodology}\label{sec:overview}

In the following, we first define the task and introduce GraphMed-LT, which consists of a patient agent (Sec.~\ref{sec:method:patient}), triplet agent (Sec.~\ref{sec:method:triplet}), patient-specific graph memory (Sec.~\ref{sec:method:graph_memory}), and a trainable doctor agent (Sec.~\ref{sec:method:doctor}). Fig.~\ref{fig:main_figure} shows how GraphMed-LT organises patient evidence across turns into graph-conditioned evidence tokens for latent clinical thought refinement in the doctor agent.

\subsection{Task Definition}

Multi-turn medical conversations simulate an iterative interaction between a doctor agent and a patient agent. The doctor agent is initially given a question, a set of candidate answers $\mathcal{A}=\{a_1,\ldots,a_C\}$, and an initial patient description $p_0$. The patient agent has access to the complete patient record $\mathcal{P}=\{p_0,p_1,\dots,p_n\}$ and reveals relevant information in response to the doctor's follow-up questions. At each turn $t$, the doctor agent either asks a follow-up question $q_t$ or outputs a final answer $\hat{a}\in\mathcal{A}$. Given $q_t$, the patient agent returns a response $r_t$ based on $\mathcal{P}$, and the observed conversation history is updated as follows:
\begin{equation}
\mathcal{C}_t
=
\mathcal{C}_{t-1}
\cup
\{(q_t,r_t)\}.
\label{eq:conversation_update}
\end{equation}
The interaction terminates when the doctor agent outputs a final answer or reaches the maximum number of turns. The goal is to select the correct answer by integrating patient evidence accumulated over multiple turns.

\subsection{The Patient Agent}\label{sec:method:patient}

The patient agent simulates the patient in a multi-turn medical conversation. Given the doctor agent's follow-up question $q_t$ and the complete patient record $\mathcal{P}$, it returns a response $r_t$ grounded in the available patient information. The patient agent does not perform diagnosis or decision-making; it only reveals patient information in response to the doctor's questions. To avoid unsupported clinical facts, the patient agent is constrained to answer using information contained in the patient record. If no relevant sentence exists for the given query, it returns a fixed response: ``\textit{The patient cannot answer this question, please do not ask this question again}'', indicating that the question cannot be answered. This constraint ensures that subsequent triplet extraction is based on dataset-supported patient evidence rather than speculative content, and also provides the basis for measuring answerable question rate. The prompt for the patient agent is provided in Appendix~\ref{app:patient_prompts}.

\subsection{Triplet Agent}\label{sec:method:triplet}

The triplet agent converts patient responses into structured clinical triplets and retrieves relevant knowledge triplets. Given the patient response $r_t$ at turn $t$, it extracts patient-specific triplets that are explicitly expressed in the response. In addition, the triplet agent retrieves relevant knowledge triplets from PrimeKG~\cite{chandak2023building}, a broad-coverage biomedical knowledge graph, using similarity matching. Specifically, the patient response and candidate knowledge triplets are encoded with the biomedical contrastive encoder BiCA~\cite{sinha2026bica}, and the top-$3$ most relevant triplets are retained as auxiliary background knowledge. The extracted and retrieved triplets are used to update the patient-specific graph memory. Details of the triplet extraction and retrieval are provided in Appendix~\ref{app:triplet_details}.

\subsection{Patient-Specific Graph Memory}
\label{sec:method:graph_memory}


The patient-specific graph memory module incrementally organises triplets acquired across conversation turns into a patient-specific graph, encodes the graph with a graph attention network (GAT)~\cite{velivckovic2017graph}, and projects the resulting representation into graph-conditioned evidence tokens for the doctor agent.

\paragraph{Graph Construction.}
The graph memory is initialised as \(\mathcal{G}_0\) using triplets extracted from the initial patient description \(p_0\). At each turn \(t\), the triplet agent provides a triplet set \(\mathcal{T}_t\). The graph memory is updated from \(\mathcal{G}_{t-1}\) to \(\mathcal{G}_t=(\mathcal{V}_t,\mathcal{E}_t)\) by adding the clinical entities and relation-labelled edges in \(\mathcal{T}_t\). Each node in \(\mathcal{V}_t\) represents a clinical entity, and each edge in \(\mathcal{E}_t\) represents a relation-labelled triplet. Source-aware edge types distinguish patient-observed triplets from retrieved knowledge. Details of graph construction are provided in Appendix~\ref{app:graph_memory_details}.

\paragraph{Graph Encoder.}
The graph encoder converts the patient-specific graph memory into a graph-level representation. Clinical entities are initialised using BiCA embeddings computed from their textual forms. We then use a GAT to propagate information over the graph and aggregate the final node representations into a graph-level representation:
\begin{equation}
\mathbf{g}_t
=
\operatorname{Pool}
\left(
\operatorname{GAT}_{\theta_{\mathrm{graph}}}
\left(
\mathcal{G}_t
\right)
\right),
\label{eq:graph_encoder}
\end{equation}
where $\theta_{\mathrm{graph}}$ denotes the trainable parameters of the graph encoder and $\operatorname{Pool}(\cdot)$ denotes attention pooling over node representations. At each turn, the GAT re-encodes the accumulated graph $\mathcal{G}_t$ rather than maintaining an incremental cache. As the graph remains compact in practice (around 36 nodes per example on iMedQA), this adds little cost relative to LLM generation. Details of the node initialisation, GAT propagation, and graph-level pooling are provided in Appendix~\ref{app:graph_memory_details}.

\paragraph{Projector.}
The graph-level representation $\mathbf{g}_t$ is mapped into the input embedding space of the doctor agent through a trainable graph-to-token projector:
\begin{equation}
\mathbf{Z}_t
=
\operatorname{Proj}_{\theta_{\mathrm{proj}}}
\left(
\mathbf{g}_t
\right)
=
[z_1,\ldots,z_m]
\in
\mathbb{R}^{m\times d},
\label{eq:graph_to_tokens}
\end{equation}
where $\theta_{\mathrm{proj}}$ denotes the trainable projector parameters, $m$ is the number of graph-conditioned evidence tokens, and $d$ is the hidden dimension of the doctor agent. These evidence tokens summarise the current patient-specific graph memory and serve as fixed evidence anchors during latent clinical thought refinement.

\subsection{Doctor Agent}
\label{sec:method:doctor}

The doctor agent is a trainable LLM for follow-up questioning and answer prediction. At turn $t$, the doctor prompt embedding $\mathbf{P}_t$ contains the question, candidate answers, observed conversation history $\mathcal{C}_{t-1}$, and task instruction. The graph-conditioned evidence tokens $\mathbf{Z}_t=[z_1,\ldots,z_m]$ produced by the patient-specific graph memory module are incorporated into the doctor agent through latent clinical thought refinement.

\paragraph{Latent Clinical Thought Refinement.}
Using graph-conditioned evidence tokens only as a static prefix may not be sufficient for the doctor agent to internalise the patient-specific graph memory, because the graph evidence is read only once before generation and may not be iteratively reconciled with the evolving clinical context. We therefore introduce latent clinical thought refinement, where graph-conditioned evidence tokens serve as fixed anchors and thought tokens serve as mutable reasoning states. Specifically, each evidence token $z_i$ is paired with one thought token $h_i^k$, providing a local refinement state for each graph-derived evidence anchor while keeping the refinement length fixed. At refinement step $k$, the input sequence is constructed as follows:
\begin{equation}
\mathbf{S}^{k}
=
[
z_1,h_1^k,
z_2,h_2^k,
\ldots,
z_m,h_m^k
].
\label{eq:interleaved_sequence}
\end{equation}
The doctor agent processes this interleaved sequence and produces hidden states at all positions. Only the hidden states at the thought-token positions are used to update the thought tokens for the next refinement step, while the evidence tokens remain unchanged within the refinement loop as graph anchors. This process is repeated for $K$ refinement steps, yielding the final graph-conditioned clinical context $\mathbf{S}^{K}$. The sequence length remains fixed during the refinement because each iteration replaces the previous thought states rather than appending new tokens.

The refined clinical context is prepended to the doctor prompt:
\begin{equation}
\mathbf{X}_t
=
[
\mathbf{S}^{K};
\mathbf{P}_t
].
\label{eq:doctor_final_input}
\end{equation}
Given $\mathbf{X}_t$, the doctor agent either asks another follow-up question or outputs a final answer. If a follow-up question is asked, the patient response is processed by the triplet agent and used to update the graph memory for the next turn. If a final answer is produced, the interaction terminates.

\begin{table*}[t]
\centering
\small
\setlength{\tabcolsep}{6.2pt}
\renewcommand{\arraystretch}{1.28}

\begin{tabular}{lcccccccccccc}
\toprule
\textbf{Method}
& \multicolumn{4}{c}{\textbf{iCRAFT-MD}}
& \multicolumn{4}{c}{\textbf{iMedQA}}
& \multicolumn{4}{c}{\textbf{iMedMCQA}} \\
\cmidrule(lr){2-5}
\cmidrule(lr){6-9}
\cmidrule(lr){10-13}
& \textbf{L8} & \textbf{Q7} & \textbf{L70} & \textbf{Q72}
& \textbf{L8} & \textbf{Q7} & \textbf{L70} & \textbf{Q72}
& \textbf{L8} & \textbf{Q7} & \textbf{L70} & \textbf{Q72} \\
\midrule

\rowcolor{lightgrayrow}
UoT
& 45.5 & 45.1 & 63.0 & 62.7
& 40.0 & 39.8 & 54.2 & 55.0
& 41.7 & 41.5 & 55.5 & 56.1 \\

CoD
& 46.2 & 45.6 & 63.8 & 63.5
& 40.7 & 40.2 & 55.0 & 55.6
& 42.4 & 42.1 & 56.2 & 56.8 \\

\rowcolor{lightgrayrow}
DDO
& 46.0 & 45.9 & 64.1 & 63.4
& 41.2 & 40.5 & 55.6 & 56.1
& 42.1 & 42.4 & 56.8 & 57.3 \\

MEDIQ
& 50.0 & 49.4 & 72.1 & 71.6
& 45.0 & 44.8 & 60.9 & 61.5
& 47.5 & 47.0 & 62.0 & 62.7 \\

\rowcolor{lightgrayrow}
MedAgentSim
& 51.8 & 50.7 & 71.8 & 72.4
& 46.6 & 46.5 & 62.8 & 63.1
& 49.2 & 48.6 & 63.8 & 64.0 \\

MEDDxAgent
& 52.7 & 52.2 & 73.7 & 72.9
& 48.6 & 47.8 & 64.1 & 65.2
& 50.4 & 49.9 & 65.4 & 65.7 \\

\rowcolor{lightgrayrow}
ProMed
& 55.2 & 54.0 & 74.6 & 74.1
& 50.3 & 49.7 & 66.8 & 67.5
& 52.0 & 51.2 & 67.9 & 68.6 \\

\midrule
\textbf{GraphMed-LT}
& \textbf{61.5$^{\dagger}$} & \textbf{60.2$^{\dagger}$} & \textbf{78.0$^{\dagger}$} & \textbf{79.1$^{\dagger}$}
& \textbf{56.4$^{\dagger}$} & \textbf{55.8$^{\dagger}$} & \textbf{72.7$^{\dagger}$} & \textbf{73.0$^{\dagger}$}
& \textbf{57.8$^{\dagger}$} & \textbf{57.0$^{\dagger}$} & \textbf{71.2$^{\dagger}$} & \textbf{72.0$^{\dagger}$} \\

\bottomrule
\end{tabular}
\caption{Accuracy (\%) comparison with baselines across datasets and doctor backbones.
L8, Q7, L70, and Q72 denote LLaMA-3-8B, Qwen-2.5-7B, LLaMA-3-70B, and Qwen-2.5-72B, respectively.
Best results are shown in \textbf{bold}.
$^{\dagger}$ indicates a significant improvement over the strongest baseline (paired t-test, $p<0.05$).
}
\label{tab:baseline_comparison}
\end{table*}

\paragraph{Training Objective.}
GraphMed-LT is trained with answer prediction supervision after graph encoding, graph-to-token projection, and latent clinical thought refinement. The trainable parameters are \(\Theta=\{\theta_{\mathrm{graph}},\theta_{\mathrm{proj}},\theta_{\mathrm{doc}}\}\), corresponding to the graph encoder, projector, and doctor agent. Latent clinical thought refinement is implemented through the hidden-state feedback procedure of the doctor agent and does not introduce a separate parameter set. Given the final input \(\mathbf{X}_t\), the model predicts a distribution \(\hat{\mathbf{y}}\) over candidate answers and is optimised with the cross-entropy loss:
\begin{equation}
\mathcal{L}_{\mathrm{ans}}(\Theta)
=
-
\sum_{c=1}^{C}
y_c\log \hat{y}_c .
\label{eq:answer_loss}
\end{equation}
The patient agent and triplet agent are frozen and are not optimised by this loss, serving only as auxiliary interaction and evidence-extraction modules. Additional details of the hidden-state refinement procedure are provided in Appendix~\ref{app:latent_thought_details}.

\section{Experimental Setup}
\label{sec:experiments}

\paragraph{Datasets}
We evaluate GraphMed-LT on three multi-turn medical conversation datasets. iMedQA~\cite{li2024mediq} is used for training and in-domain testing, with 10,178 training cases, 1,272 validation cases, and 1,273 test cases. iCRAFT-MD~\cite{johri2024craft} is a dermatology-focused zero-shot test set with 140 clinically realistic patient--doctor interactions. iMedMCQA is derived from MedMCQA~\cite{pal2022medmcqa} by converting 200 questions into multi-turn conversations, and is used as an additional zero-shot test set. Details of the datasets are provided in Appendix~\ref{app:dataset_details}.

\paragraph{Baselines}
We compare GraphMed-LT with seven multi-turn medical conversation baselines: UoT~\cite{hu2024uncertainty}, CoD~\cite{chen2025cod}, DDO~\cite{jia2025ddo}, MEDIQ~\cite{li2024mediq}, MedAgentSim~\cite{almansoori2025medagentsim}, MEDDxAgent~\cite{rose2025meddxagent}, and ProMed~\cite{ding2025promed}.All baselines use the same patient agent, candidate answer set, turn budget, stopping rule, conversation-history access, and answer-selection protocol. For DDO, MedAgentSim, and MEDDxAgent, which were not originally designed for candidate-answer selection, we preserve their information-seeking strategy and adapt only the final prediction interface to the given candidates. This keeps the interaction protocol identical while comparing how each method reasons over acquired evidence. Details are provided in Appendix~\ref{app:baseline_details}.

\paragraph{Evaluation}
We use diagnostic accuracy, measuring whether the doctor agent selects the correct answer after the multi-turn conversation. In addition, we report interaction efficiency, measured by the average number of follow-up questions asked before termination. To assess question answerability, we report the answerable question rate (AQR), defined as the proportion of follow-up questions that can be answered under the patient-record-grounded simulator protocol.

\paragraph{Implementation Details}
All experiments follow the same multi-turn medical conversation protocol. The patient agent and triplet agent are both instantiated with Qwen-2.5-72B for stable instruction-following, and kept frozen. The patient agent is shared across all methods, ensuring that every method interacts with the same patient-response mechanism. The triplet agent is used by GraphMed-LT and its ablation variants; it is fixed so that differences among these variants reflect how the same structured evidence is organised and integrated. Additional details are provided in Appendix~\ref{app:implementation_details}.

\section{Results}
\label{sec:results}

In this section, we answer five Research Questions (RQ) about GraphMed-LT: its overall effectiveness against strong baselines (Sec.~\ref{sec:rq1}), the contribution of each component (Sec.~\ref{sec:rq2}), its accuracy--efficiency trade-off in terms of interaction turns (Sec.~\ref{sec:rq3}), its robustness under different evidence settings (Sec.~\ref{sec:rq4}), and the consistency of its gains across medical specialties (Sec.~\ref{sec:rq5}).

\subsection{RQ1: How effective is GraphMed-LT for multi-turn medical conversations?}
\label{sec:rq1}

We first evaluate GraphMed-LT from two complementary perspectives: final diagnostic accuracy and question answerability. Diagnostic accuracy measures whether the doctor agent selects the correct answer after the multi-turn conversation, while AQR measures whether follow-up questions can be answered under our patient-record-grounded simulator protocol. As shown in Table~\ref{tab:baseline_comparison}, GraphMed-LT achieves the best accuracy in all twelve dataset--backbone settings. Compared with the strongest baseline in each setting, GraphMed-LT provides consistent gains of 3.3--6.3 percentage points. For example, on iCRAFT-MD, GraphMed-LT improves accuracy over ProMed from 55.2 to 61.5 with LLaMA-3-8B and from 74.1 to 79.1 with Qwen-2.5-72B. On iMedQA, GraphMed-LT reaches 73.0 with Qwen-2.5-72B, compared with 67.5 for ProMed. On iMedMCQA, GraphMed-LT also achieves the highest accuracy, reaching 72.0 with Qwen-2.5-72B compared with 68.6 for the strongest baseline.

\begin{figure}[t]
    \centering
    \includegraphics[width=\linewidth]{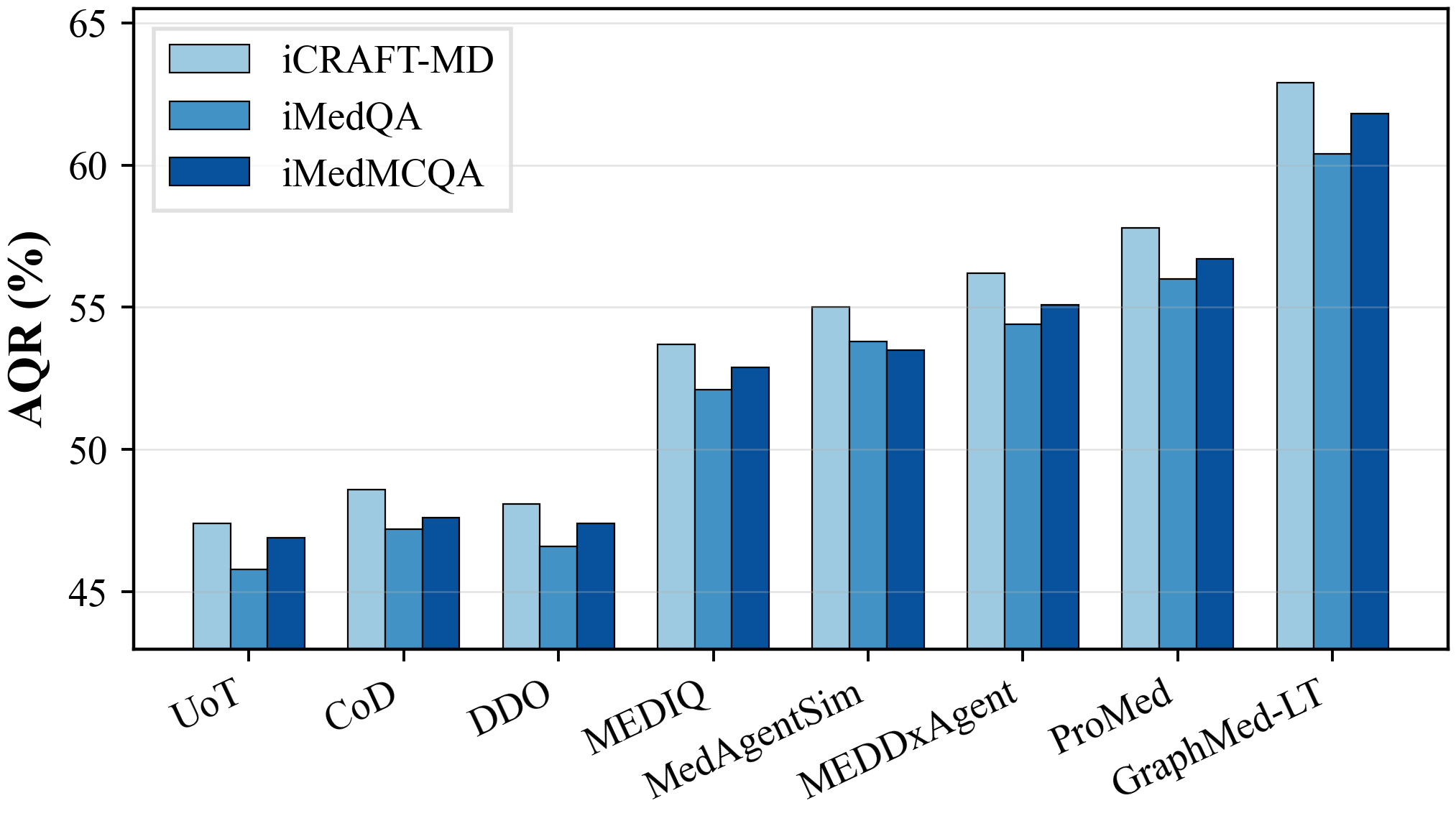}
    \caption{Answerable question rate (\%) comparison with multi-turn medical conversation baselines across datasets.
    Qwen-2.5-72B is used as the doctor backbone. 
    }
    \label{fig:aqr_qwen72}
\end{figure}

Fig.~\ref{fig:aqr_qwen72} further reports AQR using Qwen-2.5-72B as a representative strong doctor backbone, while Table~\ref{tab:baseline_comparison} shows that the accuracy trend is consistent across all evaluated backbones. GraphMed-LT obtains 62.9 AQR on iCRAFT-MD, 60.4 on iMedQA, and 61.8 on iMedMCQA, improving over the strongest baseline by 5.1, 4.4, and 5.1 percentage points, respectively. This indicates that GraphMed-LT asks follow-up questions that are more often answerable under the benchmark protocol, rather than improving accuracy through indiscriminate information seeking. Together, the accuracy and AQR results across the three datasets suggest that GraphMed-LT improves both final answer selection and protocol-level information seeking in multi-turn medical conversations.

\subsection{RQ2: How does each component affect GraphMed-LT?}
\label{sec:rq2}

\begin{table*}[!ht]
\centering
\small
\setlength{\tabcolsep}{7pt}
\renewcommand{\arraystretch}{1.24}

\begin{tabular}{lcccccccccccc}
\toprule
\textbf{Method}
& \multicolumn{4}{c}{\textbf{iCRAFT-MD}}
& \multicolumn{4}{c}{\textbf{iMedQA}}
& \multicolumn{4}{c}{\textbf{iMedMCQA}} \\
\cmidrule(lr){2-5}
\cmidrule(lr){6-9}
\cmidrule(lr){10-13}
& \textbf{L8} & \textbf{Q7} & \textbf{L70} & \textbf{Q72}
& \textbf{L8} & \textbf{Q7} & \textbf{L70} & \textbf{Q72}
& \textbf{L8} & \textbf{Q7} & \textbf{L70} & \textbf{Q72} \\
\midrule

\rowcolor{lightgrayrow}
Conversation History
& 50.0 & 49.4 & 72.1 & 71.6
& 45.0 & 44.8 & 60.9 & 61.5
& 47.5 & 47.0 & 62.0 & 62.7 \\

Triplets as Text
& 53.7 & 53.1 & 72.4 & 72.6
& 49.8 & 48.9 & 63.2 & 64.5
& 51.0 & 50.6 & 62.9 & 63.6 \\

\rowcolor{lightgrayrow}
w/o Retrieved Triplets
& 60.6 & 59.5 & 77.1 & 78.2
& 55.5 & 55.1 & 71.8 & 72.3
& 56.9 & 56.3 & 70.3 & 71.1 \\

w/o Graph Memory
& 57.4 & 56.0 & 75.2 & 76.0
& 52.1 & 51.6 & 68.4 & 69.2
& 53.8 & 53.1 & 67.0 & 67.8 \\

\rowcolor{lightgrayrow}
w/o Latent Thoughts
& 55.8 & 54.6 & 73.8 & 74.5
& 50.9 & 50.2 & 66.0 & 66.8
& 52.4 & 51.7 & 65.3 & 66.0 \\

\midrule
\textbf{GraphMed-LT}
& \textbf{61.5} & \textbf{60.2} & \textbf{78.0} & \textbf{79.1}
& \textbf{56.4} & \textbf{55.8} & \textbf{72.7} & \textbf{73.0}
& \textbf{57.8} & \textbf{57.0} & \textbf{71.2} & \textbf{72.0} \\

\bottomrule
\end{tabular}
\caption{Ablation study across datasets and doctor backbones.
L8, Q7, L70, and Q72 denote LLaMA-3-8B, Qwen-2.5-7B, LLaMA-3-70B, and Qwen-2.5-72B, respectively.
Best results are shown in \textbf{bold}.
}
\label{tab:component_ablation}
\end{table*}

\begin{figure*}[!ht]
    \centering
    \includegraphics[width=\linewidth]{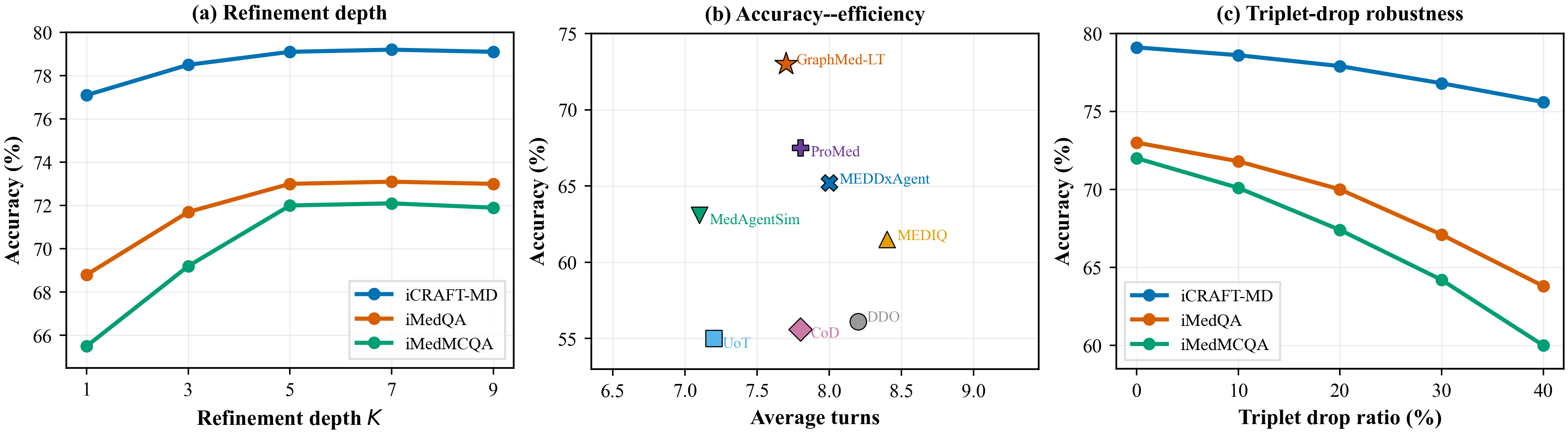}
    \caption{
    \textbf{(a)} Accuracy under different latent refinement depths.
    \textbf{(b)} Accuracy--efficiency trade-off on iMedQA.
    \textbf{(c)} Robustness to incomplete triplet extraction.
    }
    \label{fig:analysis}
\end{figure*}

We conduct comprehensive component ablations to examine whether GraphMed-LT's gains come from structured triplet preprocessing, patient-specific graph memory, retrieved knowledge, or latent clinical thought refinement. To further isolate upstream triplet preprocessing from downstream evidence integration, all triplet-based variants reuse the same extracted triplets and differ only in how they are organised and injected into the doctor agent; in particular, \textit{w/o Graph Memory} keeps the same extracted and retrieved triplets but removes graph construction and GAT propagation, aggregating triplet representations into a non-graph evidence representation before projection. As shown in Table~\ref{tab:component_ablation}, \textit{Conversation History}, which provides the accumulated dialogue history directly to the doctor agent without triplet extraction or graph memory, performs weakest in most settings, while \textit{Triplets as Text} improves over it, suggesting that structured clinical facts provide a clearer evidence interface than raw dialogue history. 

Removing latent thoughts causes the largest accuracy drop, from 73.0 to 66.8 on iMedQA with Qwen-2.5-72B, showing that hidden-state refinement is central to internalising graph-conditioned evidence. Removing graph memory leads to the second-largest accuracy drop, from 73.0 to 69.2, indicating that graph-based organisation connects patient evidence across turns. Removing the retrieved triplets causes a smaller accuracy decrease, from 73.0 to 72.3, suggesting that retrieved knowledge is useful but not the primary source of improvement. Overall, these ablations show that GraphMed-LT's gains are not solely due to triplet preprocessing; the main improvements come from latent thought refinement and patient-specific graph memory.

\subsection{RQ3: Does GraphMed-LT improve the accuracy--efficiency trade-off?}
\label{sec:rq3}
We further examine whether GraphMed-LT improves diagnostic accuracy without substantially increasing the number of interaction turns. We conduct this analysis on iMedQA with Qwen-2.5-72B as a representative setting, and report accuracy against the average number of follow-up questions. As shown in Fig.~\ref{fig:analysis}(b), GraphMed-LT achieves 73.0 accuracy with 7.7 average turns, compared with 67.5 for ProMed at 7.8 turns and 65.2 for MEDDxAgent at 8.0 turns. Although MedAgentSim uses slightly fewer turns, its accuracy is substantially lower than GraphMed-LT. These results show that GraphMed-LT achieves higher accuracy than the strongest baselines while using a comparable number of follow-up questions. This suggests that our proposed approach does not necessarily rely on longer conversations; instead, patient-specific graph memory can help the doctor agent use acquired evidence more effectively within a compact interaction process. See Appendix~\ref{app:efficiency} for a detailed runtime comparison.

\subsection{RQ4: Is GraphMed-LT stable under different evidence settings?}
\label{sec:rq4}
We analyse GraphMed-LT's sensitivity to latent refinement depth and simulated triplet missingness, where extracted triplets are randomly dropped. As shown in Fig.~\ref{fig:analysis}(a), increasing the number of latent refinement steps $K$ improves accuracy and then saturates: iMedQA rises from 68.8 at $K=1$ to 73.0 at $K=5$, iMedMCQA from 65.5 to 72.0, and iCRAFT-MD from 77.1 to 79.1, with little gain beyond $K=5$. Fig.~\ref{fig:analysis}(c) shows that accuracy decreases as more triplets are dropped. With 20\% of triplets removed, GraphMed-LT still reaches 77.9 on iCRAFT-MD, 70.0 on iMedQA, and 67.4 on iMedMCQA. These results support $K=5$ as a stable and efficient choice and suggest that GraphMed-LT is robust to moderate triplet missingness. Appendix~\ref{app:noise} further evaluates robustness under structured
triplet noise beyond random dropping.

\begin{figure*}[!ht]
    \centering
    \includegraphics[width=\linewidth]{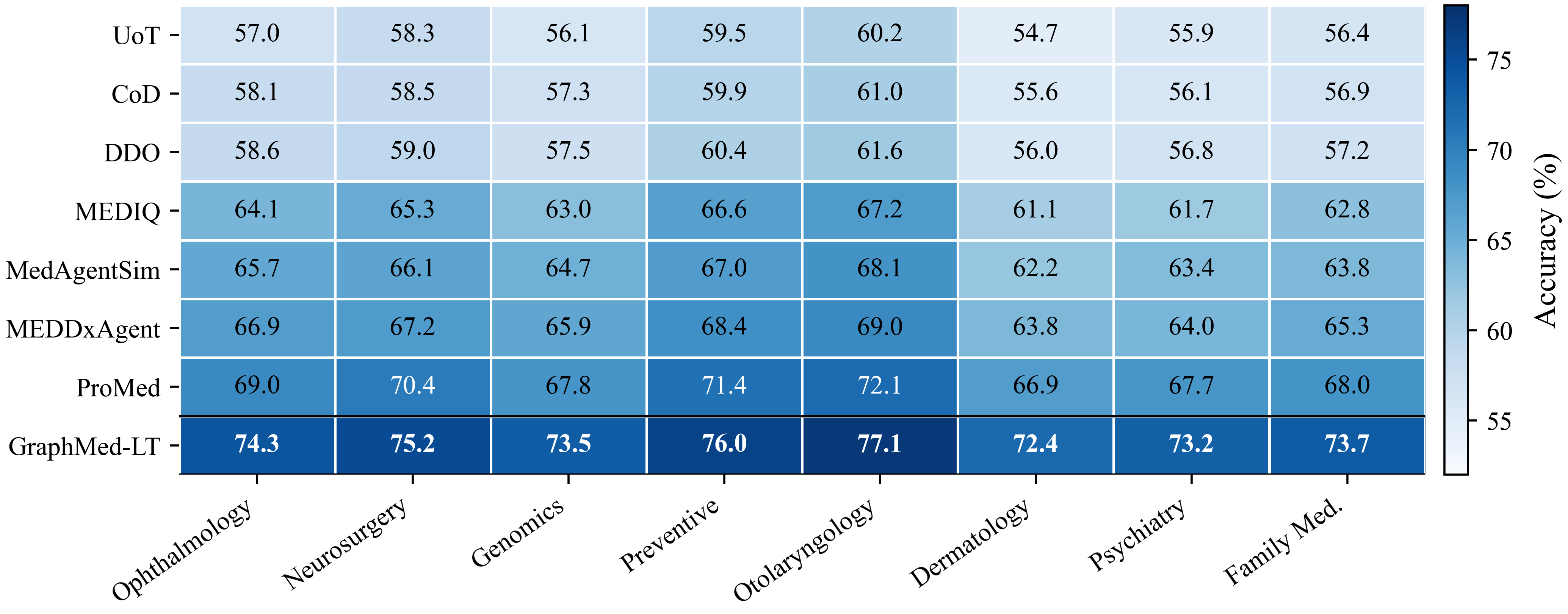}
    \caption{
    Accuracy (\%) comparison across eight medical specialty groups on iMedQA.
    }
    \label{fig:specialty_heatmap}
\end{figure*}

\subsection{RQ5: Are GraphMed-LT's gains consistent across medical specialties?}
\label{sec:rq5}
We further examine whether GraphMed-LT's gains are consistent across different clinical domains, represented by medical specialty groups within the benchmark setting. As shown in Fig.~\ref{fig:specialty_heatmap}, GraphMed-LT achieves the highest accuracy across all eight specialties, ranging from 72.4 on Dermatology to 77.1 on Otolaryngology. Compared with representative baselines, GraphMed-LT improves consistently across specialties. For example, it improves over MEDDxAgent by 7.6 points on Genomics and 8.6 points on Dermatology, and over ProMed by 5.7 and 5.5 points on the same specialties. On average, GraphMed-LT improves over ProMed, the strongest overall baseline in Table~\ref{tab:baseline_comparison}, from 69.2 to 74.4, suggesting that its gains are not confined to a single specialty group.

\section{Related Work}
In the following, we review two lines of related work, multi-turn medical conversation agents and knowledge-integrated medical models, and position our contributions in the existing literature.

\subsection{Multi-Turn Medical Conversation Agents}
Multi-turn medical conversations require a doctor agent to collect patient information across dialogue turns and make diagnostic decisions from progressively acquired evidence~\cite{rezaei-etal-2025-agentic,zhu2025ask,wang2025medkgi}. Early automatic diagnosis systems formulate this process as sequential decision-making, where the agent asks about additional symptoms before predicting the final disease. Recent generation-based models further improve symptom inquiry and diagnosis through sequence modelling and attention mechanisms. Diaformer~\cite{chen2022diaformer} formulated automatic diagnosis as symptom sequence generation, MTDiag~\cite{hou2023mtdiag} jointly modelled symptom inquiry and disease prediction, and HAIformer~\cite{zhao2024haiformer} incorporated human--AI collaboration for diagnostic decision-making. More recent LLM-based agents have improved active information seeking and diagnostic reasoning in multi-turn settings~\cite{zhou2026note2chat,feng2026doctoragent}. These methods show the value of question-asking medical agents, but patient evidence is still largely maintained as dialogue history for multi-turn reasoning. In contrast, GraphMed-LT explicitly organises and refines cross-turn patient evidence into a graph memory that conditions subsequent doctor-agent reasoning.

\subsection{Knowledge-Integrated Medical Models}
Knowledge integration has become an important direction for improving medical QA, especially when answers require external evidence or multi-step reasoning~\cite{varshney2023knowledge,yu2025medkgeval,meng2024heterogeneous}. MedRAG~\cite{xiong2024benchmarking} evaluated how medical corpora, retrievers, and LLM backbones affect evidence-enhanced medical QA. RAG$^2$~\cite{sohn2025rationale} used model-generated rationales to guide biomedical evidence retrieval and filter noisy snippets. MedGraphRAG~\cite{wu2025medical} organised medical documents into graph structures for evidence-grounded generation, while RGAR~\cite{liang-etal-2025-rgar} incorporated patient-specific factual knowledge from electronic health records. Although these studies have achieved strong performance by retrieving external medical evidence, they mainly organise evidence for a given query. GraphMed-LT instead updates a patient-specific graph memory throughout the conversation and refines graph-conditioned evidence inside the doctor agent.

\section{Conclusions}
\label{sec:conclusion}

Multi-turn medical conversations require doctor agents to integrate patient evidence progressively acquired across dialogue turns. Existing methods often rely on accumulated conversation histories, leaving clinical evidence fragmented for multi-turn reasoning. We proposed GraphMed-LT, a novel patient-specific graph memory approach with latent clinical thought refinement. GraphMed-LT extracts patient-specific clinical triplets, retrieves auxiliary knowledge triplets, organises them into an incrementally updated graph memory, and refines graph-conditioned evidence tokens inside a trainable doctor agent through hidden-state feedback. Experiments on three multi-turn medical datasets show that GraphMed-LT consistently outperforms strong baselines across multiple LLM backbones, achieving up to a 6.3 percentage-point absolute improvement. Further analyses showed that GraphMed-LT asks more answerable follow-up questions, improves the accuracy--efficiency trade-off and provides consistent gains across medical specialties. These findings suggest that patient-specific graph memory with latent clinical thought refinement provides an effective mechanism for reliable multi-turn medical reasoning, thereby supporting more robust multi-turn medical conversations.

\section*{Limitations}
\label{sec:limitations}

This work has several limitations. (1) Our evaluation focuses on benchmark-style multi-turn medical conversations with candidate answers, which enables a controlled and reproducible comparison. Future work can extend GraphMed-LT to open-ended diagnostic generation, longitudinal patient histories, and clinician-authored conversations. (2) GraphMed-LT is evaluated with patient simulators constructed from structured patient records. While this setting ensures that patient responses are grounded in dataset-supported evidence, such controlled simulation is known to trade naturalness and behavioural diversity for controllability~\cite{tennenholtz2026controllable}. (3) Our current graph memory uses extracted patient-specific triplets and retrieved background triplets under a predefined relation schema. Future work can enrich this schema with broader clinical ontologies, uncertainty-aware extraction, and clinician-validated knowledge sources, further improving graph construction and evidence grounding in real-world medical settings.

\section*{Acknowledgements}
The authors acknowledge the use of resources provided by the Isambard-AI National AI Research Resource (AIRR)~\cite{mcintosh2024isambard}. Isambard-AI is operated by the University of Bristol and is funded by the UK Government's Department for Science, Innovation and Technology (DSIT) via UK Research and Innovation; and the Science and Technology Facilities Council [ST/AIRR/I-A-I/1023].

K.Y. acknowledges support from Cancer Research UK (EDDPGM-Nov21/100001, DRCMDP-Nov23/100010 and core funding to the CRUK Scotland Institute (A31287)), BBSRC BB/V016067/1, Prostate Cancer UK MA-TIA22-001 and EU Horizon 2020 grant ID: 101016851.

\bibliography{custom}
\clearpage
\appendix
\section{Appendix}\label{sec:appendix}

\subsection{Computing Hardware}\label{app:model_details}

All experiments were conducted on GPU nodes equipped with 8 NVIDIA GH200 GPUs. These resources provide sufficient memory and computational capacity to support large-scale model inference and training under the experimental settings considered in this work.

\subsection{Implementation Details}
\label{app:implementation_details}

\begin{figure}[!ht]
    \centering
    \includegraphics[width=\linewidth]{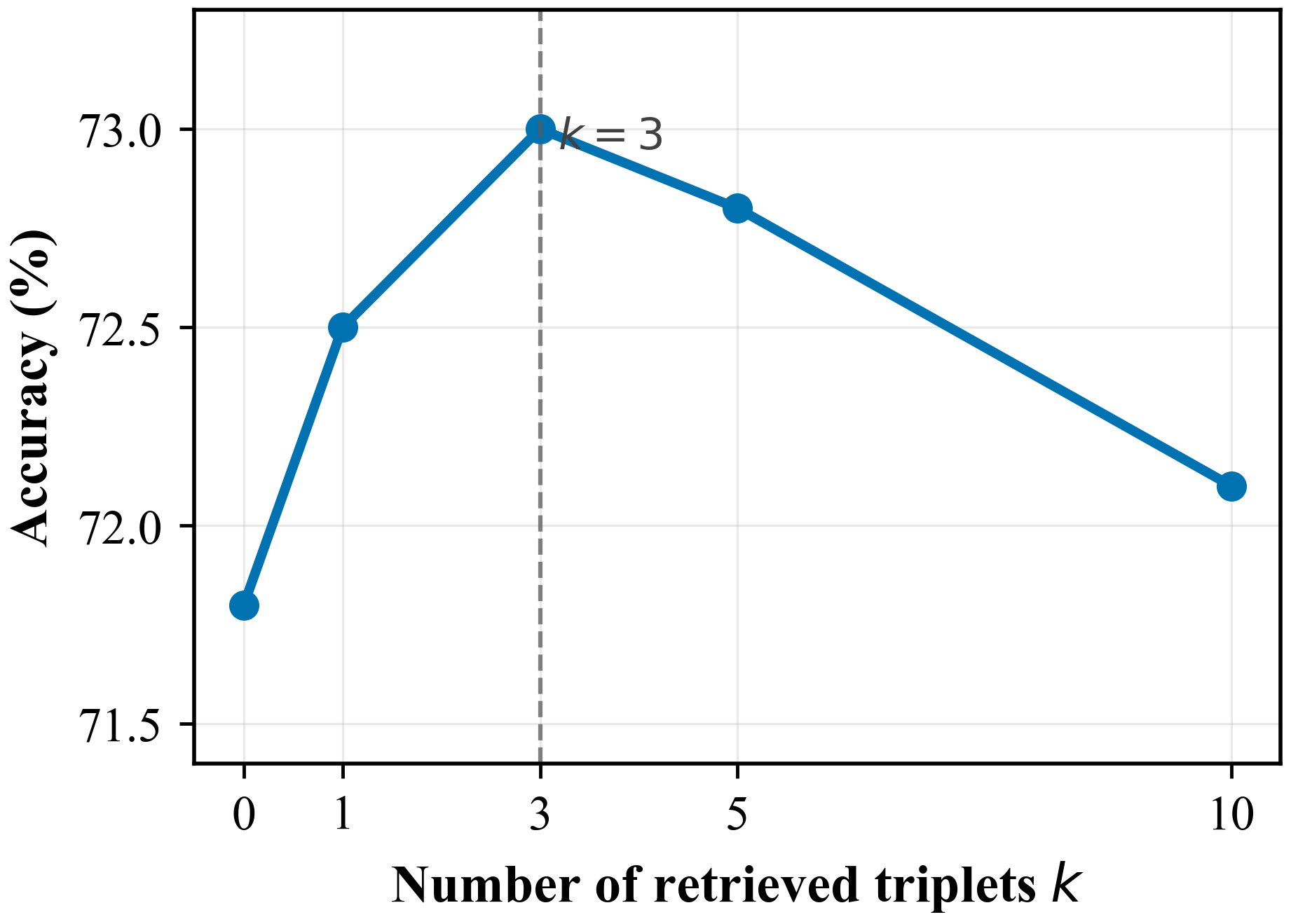}
    \caption{
    \textbf{Effect of retrieved knowledge triplets.}
    We vary the number of retrieved knowledge triplets $k$ on iMedQA using Qwen-2.5-72B as the doctor backbone.
    Accuracy improves when a small number of knowledge triplets is retrieved, peaks at $k=3$, and decreases when more triplets are added, suggesting that excessive retrieval can introduce irrelevant background knowledge.
    }
    \label{fig:appendix_topk_retrieval}
\end{figure}

\begin{figure}[!ht]
    \centering
    \includegraphics[width=\linewidth]{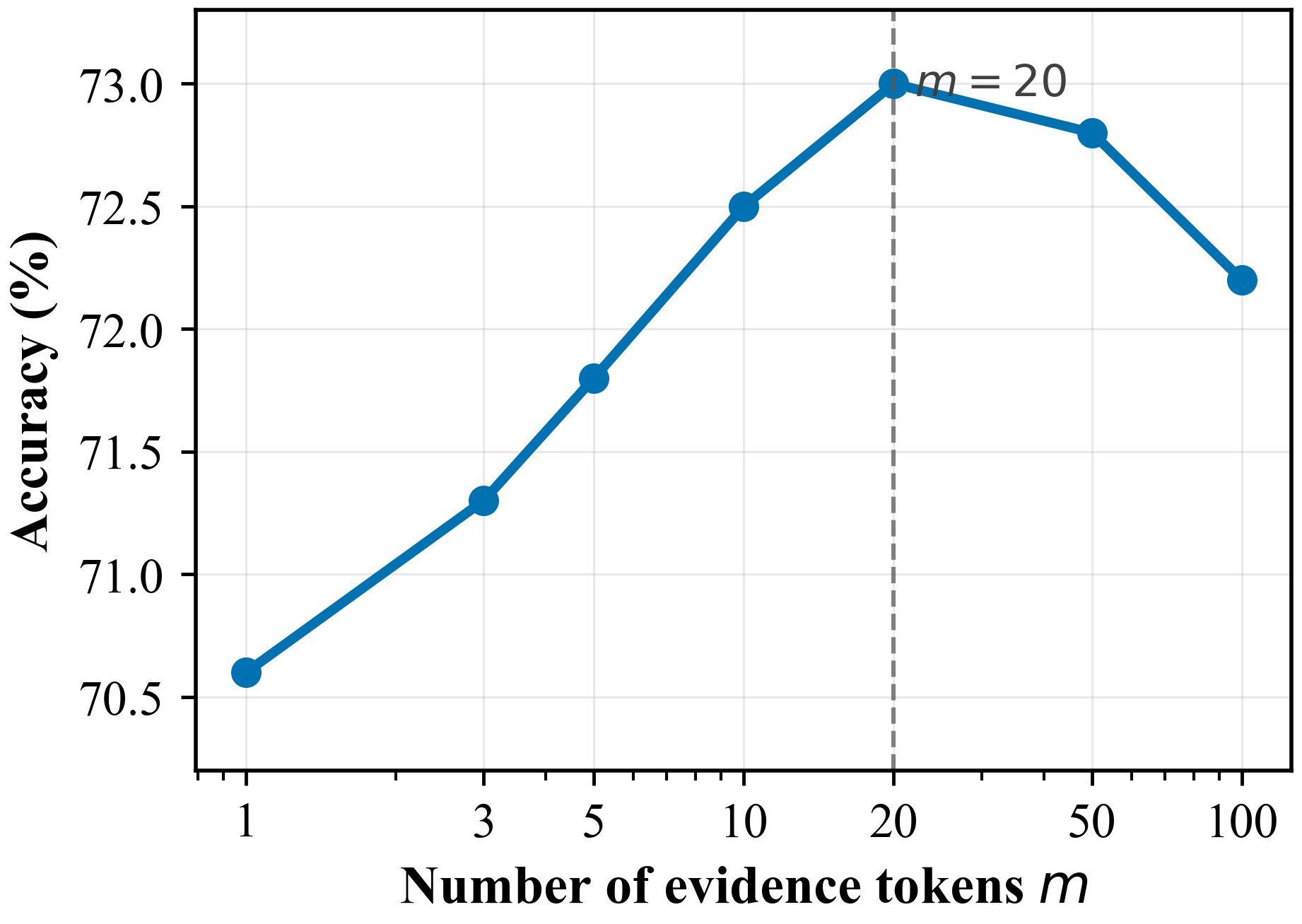}
    \caption{
    \textbf{Effect of graph-conditioned evidence-token budget.}
    We vary the number of graph-conditioned evidence tokens $m$ on iMedQA using Qwen-2.5-72B as the doctor backbone.
    Accuracy increases with larger evidence-token budgets and peaks at $m=20$, while larger budgets provide no further gain.
    }
    \label{fig:appendix_evidence_token_budget}
\end{figure}

\paragraph{Hyperparameter sensitivity.}
We further vary the number of retrieved triplets and the number of graph-conditioned evidence tokens. As shown in Fig.~\ref{fig:appendix_topk_retrieval} and Fig.~\ref{fig:appendix_evidence_token_budget}, top-$3$ retrieval and $m=20$ evidence tokens provide a strong and compact setting for the main experiments. We select hyperparameters using iMedQA development-set accuracy and keep the selected settings fixed for iMedQA test, iCRAFT-MD, and iMedMCQA. During evaluation, all methods use the same decoding setting, with temperature $0.2$, top-$p=0.9$.

\paragraph{Follow-up question generation and stopping.}
For all methods, the doctor agent generates follow-up questions under the same prompt format and decoding setting. We do not use a predefined question pool; instead, each method generates its own follow-up questions during the conversation. A conversation stops once the doctor agent gives a final answer or reaches the maximum budget of 10 turns. The final answer is always selected from the given candidate answers.

For GraphMed-LT, the triplet agent retrieves the top-$3$ knowledge triplets at each turn. The graph-to-token projector produces $m=20$ graph-conditioned evidence tokens from the current patient-specific graph. Each evidence token is paired with one latent thought token, and the refinement loop is run for $K=5$ steps unless otherwise specified. Table~\ref{tab:hyperparameter_settings} lists the full setting.

\begin{table}[!ht]
\centering
\small
\setlength{\tabcolsep}{4pt}
\renewcommand{\arraystretch}{1.15}
\begin{tabular}{ll}
\toprule
\textbf{Hyperparameter} & \textbf{Value} \\
\midrule
Maximum interaction turns & 10 \\
Maximum generation tokens & 256 per response \\
Retrieved knowledge triplets & 3 \\
Graph encoder & GAT \\
GAT layers & 2 \\
GAT hidden dimension & 256 \\
GAT attention heads & 4 \\
Graph pooling & Attention pooling \\
Projector type & Two-layer MLP \\
Projector activation & SiLU \\
Evidence tokens $m$ & 20 \\
Refinement steps $K$ & 5 \\
Optimiser & AdamW \\
Learning rate & $1\times10^{-5}$ \\
Weight decay & 0.01 \\
Batch size & 64 \\
Training epochs & 5 \\
Evaluation temperature & 0.2 \\
Evaluation top-$p$ & 0.9 \\
Doctor-agent tuning & Full fine-tuning \\
Patient agent & Frozen \\
Triplet agent & Frozen \\
Patient/triplet agent backbone & Qwen-2.5-72B \\
\bottomrule
\end{tabular}
\caption{Hyperparameter settings for GraphMed-LT.}
\label{tab:hyperparameter_settings}
\end{table}

\paragraph{Frozen agent configuration and quality checks.}
\label{app:frozen_agent_quality}

GraphMed-LT uses two frozen auxiliary agents. The patient agent is shared across all methods and answers follow-up questions only from atomic facts derived from the complete patient record; unsupported questions trigger a fixed unanswerable response. This keeps patient-response generation identical across methods and prevents unsupported clinical information from entering the conversation. The triplet agent is used only by GraphMed-LT and its ablation variants. It extracts patient-specific triplets from the current patient response and retrieves auxiliary knowledge triplets from PrimeKG using BiCA-based similarity, without performing answer prediction or diagnostic reasoning.

\begin{table}[t]
\centering
\small
\setlength{\tabcolsep}{5pt}
\renewcommand{\arraystretch}{1.12}
\begin{tabular}{lc}
\toprule
\textbf{Quality check} & \textbf{Rate (\%)} \\
\midrule
Extracted triplets supported by patient response & 98.0 \\
Retrieved triplets clinically relevant to response & 78.0 \\
Unanswerable responses correctly triggered & 89.0 \\
\bottomrule
\end{tabular}
\caption{
Manual quality check of the frozen auxiliary agents on 100 randomly sampled iMedQA development examples.
}
\label{tab:auxiliary_quality}
\end{table}

We manually inspect 100 randomly sampled iMedQA development examples to estimate the quality of the frozen auxiliary agents. As shown in Table~\ref{tab:auxiliary_quality}, 98.0\% of extracted triplets are supported by the patient response, 78.0\% of retrieved triplets are clinically relevant to the response, and 89.0\% of unanswerable cases correctly trigger the fixed unanswerable response. These results suggest that the auxiliary agents provide reliable but non-oracle upstream evidence.

\paragraph{Potential risks and intended use.}
Although GraphMed-LT is evaluated only on benchmark-style multi-turn medical conversation tasks, incorrect diagnoses, unsupported follow-up questions, or incomplete patient-evidence integration could be harmful if the system were used directly in biomedical or clinical decision-making. GraphMed-LT should therefore be treated as a research system for evaluating evidence-grounded multi-turn medical reasoning, rather than as a tool for clinical diagnosis, treatment recommendation, triage, or patient-facing deployment.

\paragraph{Artifact licences and intended use.}
We use publicly available datasets, models, and knowledge resources in accordance with their stated research-use terms and licences. iCRAFT-MD, iMedQA, iMedMCQA, MedMCQA, PrimeKG, BiCA, LLaMA, Qwen, and the compared baseline systems are used only for research evaluation. The derived multi-turn conversation settings and patient-specific graph memories are constructed for benchmark evaluation and are not intended for clinical deployment or patient-facing use. The released GraphMed-LT code will be distributed for research.

\subsection{Triplet Retrieval Details}
\label{app:triplet_details}

This section provides additional details for retrieving auxiliary knowledge triplets. At turn $t$, the triplet agent extracts patient-specific triplets from the patient response and retrieves relevant knowledge triplets from PrimeKG~\cite{chandak2023building}. The extracted triplets represent patient-observed evidence, while the retrieved triplets provide auxiliary background knowledge. Each candidate triplet $\tau=(h,\rho,o)$ is linearised into a textual sequence containing its head entity, relation, and tail entity. We encode both the patient response and each linearised candidate triplet using BiCA, and compute their cosine similarity as the relevance score. The top-$3$ triplets with the highest relevance scores are retained as retrieved knowledge triplets.

\subsection{Patient-Specific Graph Memory Details}
\label{app:graph_memory_details}

This section provides implementation details for the patient-specific graph memory module. We initialise $\mathcal{G}_0$ by extracting triplets from the initial patient description $p_0$. At turn $t$, the graph memory is updated using patient-specific triplets extracted from the patient response and relevant knowledge triplets retrieved from the external corpus. Each clinical entity is represented as a node, and each triplet $(u,\rho,v)$ is represented as a directed relation-labelled edge from $u$ to $v$.

\paragraph{Source-aware edge typing.}
Extracted and retrieved triplets are assigned different edge source types:
\begin{equation}
\sigma(e)
\in
\{\mathrm{patient},\mathrm{retrieved}\}.
\label{eq:source_edge_type}
\end{equation}
Edges with $\sigma(e)=\mathrm{patient}$ are extracted from patient responses, while edges with $\sigma(e)=\mathrm{retrieved}$ are retrieved from the external triplet corpus. This preserves the distinction between patient-observed evidence and auxiliary background knowledge. The relation label and source type are represented by trainable embeddings and used in the graph encoder attention score.

\paragraph{Node initialisation.}
Each clinical entity is represented using a BiCA embedding computed from its textual form:
\begin{equation}
\mathbf{h}_i^{(0)}
=
\operatorname{BiCA}(\operatorname{Text}(v_i)),
\label{eq:node_initialisation}
\end{equation}
where $\operatorname{Text}(v_i)$ denotes the textual form of node $v_i$.

\paragraph{GAT-based graph encoder.}
The graph encoder is implemented using a graph attention network (GAT). At layer $\ell$, node $i$ attends to its neighbours $\mathcal{N}(i)$ and updates its representation as:
\begin{equation}
\mathbf{h}_{i}^{(\ell+1)}
=
\sigma
\left(
\sum_{j\in\mathcal{N}(i)}
\alpha_{ij}^{(\ell)}
\mathbf{W}^{(\ell)}
\mathbf{h}_{j}^{(\ell)}
\right),
\label{eq:gat_update}
\end{equation}
where $\mathbf{W}^{(\ell)}$ is a trainable transformation matrix, $\alpha_{ij}^{(\ell)}$ is the attention weight from node $i$ to node $j$, and $\sigma(\cdot)$ is a non-linear activation function. The attention weight $\alpha_{ij}^{(\ell)}$ is computed from the transformed node representations together with trainable embeddings of the relation label and source type of the edge between nodes $i$ and $j$.

\paragraph{Graph-to-token projector.}
After graph encoding and attention pooling, the graph-level representation $\mathbf{g}_t$ is mapped into graph-conditioned evidence tokens through a trainable projector:
\begin{equation}
\begin{aligned}
\mathbf{Z}_t
&=
\operatorname{Reshape}
\left(
\operatorname{MLP}_{\theta_{\mathrm{proj}}}(\mathbf{g}_t)
\right) \\
&=
[z_1,\ldots,z_m]
\in
\mathbb{R}^{m\times d},
\end{aligned}
\label{eq:projector}
\end{equation}
where $\theta_{\mathrm{proj}}$ denotes the projector parameters, $m$ is the number of graph-conditioned evidence tokens, and $d$ is the hidden dimension of the doctor agent.

\begin{table*}[t]
\centering
\small
\caption{Dataset statistics. iMedQA provides train/dev/test splits, while iCRAFT-MD and iMedMCQA are used as test-only evaluation sets.}
\label{tab:dataset_statistics_appendix}
\setlength{\tabcolsep}{5pt}
\renewcommand{\arraystretch}{1.15}
\begin{tabular}{lcccc}
\toprule
\textbf{Dataset} & \textbf{Train} & \textbf{Dev} & \textbf{Test} & \textbf{Source} \\
\midrule
iMedQA    & 10,178 & 1,272 & 1,273 & MedQA~\cite{jin2021disease} \\
iCRAFT-MD & --     & --    & 140   & CRAFT-MD~\cite{johri2023testing,johri2024craft} \\
iMedMCQA  & --     & --    & 200   & MedMCQA~\cite{pal2022medmcqa} \\
\bottomrule
\end{tabular}
\end{table*}

\subsection{Latent Clinical Thought Refinement Details}
\label{app:latent_thought_details}

This section provides implementation details for latent clinical thought refinement inside the doctor agent. The refinement procedure keeps graph-conditioned evidence tokens fixed and iteratively updates their corresponding thought tokens in continuous hidden space.

\paragraph{Thought initialisation.}
Given graph-conditioned evidence tokens $\mathbf{Z}=[z_1,\ldots,z_m]\in\mathbb{R}^{m\times d}$, we initialise one thought token for each evidence token. In our implementation, the initial thought states are obtained by applying the doctor agent to the evidence tokens and reading the corresponding hidden states:
\begin{equation}
\mathbf{H}^{0}
=
[h_1^0,\ldots,h_m^0]
=
\operatorname{Doctor}_{\theta_{\mathrm{doc}}}(\mathbf{Z})_{1:m}.
\label{eq:thought_initialisation}
\end{equation}

\paragraph{Interleaved hidden-state refinement.}
At refinement step $k$, each fixed evidence token is interleaved with its corresponding thought token:
\begin{equation}
\mathbf{S}^{k}
=
[
z_1,h_1^k,
z_2,h_2^k,
\ldots,
z_m,h_m^k
]
\in \mathbb{R}^{2m\times d}.
\label{eq:appendix_interleaved_sequence}
\end{equation}
The doctor agent processes this interleaved sequence and returns hidden states for all $2m$ positions:
\begin{equation}
\begin{aligned}
\widetilde{\mathbf{H}}^{k+1}
&=
[
\widetilde{\mathbf{H}}^{k+1}_{1},
\widetilde{\mathbf{H}}^{k+1}_{2},
\ldots,
\widetilde{\mathbf{H}}^{k+1}_{2m}
] \\
&=
\operatorname{Doctor}_{\theta_{\mathrm{doc}}}
(\mathbf{S}^{k}).
\end{aligned}
\label{eq:appendix_hidden_update}
\end{equation}
Because the odd positions correspond to evidence tokens and the even positions correspond to thought tokens, only the even-position hidden states are used to form the next-step thought states:
\begin{equation}
h_i^{k+1}
=
\widetilde{\mathbf{H}}^{k+1}_{2i},
\qquad
i=1,\ldots,m,
\label{eq:appendix_thought_update_single}
\end{equation}
or equivalently,
\begin{equation}
\mathbf{H}^{k+1}
=
[
\widetilde{\mathbf{H}}^{k+1}_{2},
\widetilde{\mathbf{H}}^{k+1}_{4},
\ldots,
\widetilde{\mathbf{H}}^{k+1}_{2m}
].
\label{eq:appendix_thought_update}
\end{equation}
The graph-conditioned evidence tokens $z_i$ are re-used unchanged at every refinement step, while only the thought tokens are updated:
\begin{equation}
\mathbf{S}^{k+1}
=
[
z_1,h_1^{k+1},
z_2,h_2^{k+1},
\ldots,
z_m,h_m^{k+1}
].
\label{eq:appendix_next_sequence}
\end{equation}
Thus, the refinement length remains fixed at $2m$ and does not grow with the number of refinement steps.

\subsection{Datasets Description}
\label{app:dataset_details}

\paragraph{iMedQA.}
iMedQA is reformulated from MedQA~\cite{jin2021disease} into a multi-turn medical conversation setting. Each case contains an initial patient description, a complete patient record accessible only to the patient agent, an MCQ, candidate answers, and the gold answer. The doctor agent initially observes only partial patient information and must ask follow-up questions to acquire additional evidence before making a final answer prediction.

\paragraph{iCRAFT-MD.}
iCRAFT-MD is a dermatology-focused multi-turn medical conversation dataset~\cite{johri2023testing,johri2024craft}. It is used as an external evaluation set to assess whether the model generalises to clinically realistic patient--doctor interactions beyond the main training benchmark.

\paragraph{iMedMCQA.}
We construct iMedMCQA from MedMCQA~\cite{pal2022medmcqa} as an additional test-only dataset. We sample 200 medical MCQs and convert them into multi-turn medical conversations. For each case, the original answer option is preserved as the gold label. The question stem is decomposed into patient-specific atomic facts. Only partial information, such as demographic information and chief complaint when available, is exposed as the initial patient description, while the remaining facts are hidden from the doctor agent and can only be obtained through follow-up questions.

\subsection{Descriptions of Baselines}
\label{app:baseline_details}

We compare GraphMed-LT with baselines that support follow-up questioning and maintain information acquired across turns.

\paragraph{MEDIQ.}
MEDIQ~\cite{li2024mediq} formulates medical QA as an interactive doctor--patient reasoning process. The doctor agent asks follow-up questions to a patient simulator and reasons over accumulated conversation history before producing a final answer.

\paragraph{ProMed.}
ProMed~\cite{ding2025promed} encourages proactive information seeking using reinforcement learning with Shapley Information Gain rewards. We adapt it to the same MCQ-based multi-turn protocol by allowing the doctor agent to use acquired patient responses before selecting the final answer.

\paragraph{DDO.}
DDO~\cite{jia2025ddo} formulates medical conversation as a dual-decision optimisation problem, jointly modelling inquiry and diagnosis. Since DDO was originally designed for symptom-centric disease diagnosis, we adapt its inquiry-and-diagnosis process to MCQ answer selection under the same interaction protocol.

\paragraph{UoT.}
UoT~\cite{hu2024uncertainty} uses uncertainty-aware planning to guide information seeking. In our setting, it selects follow-up questions based on uncertainty in the current reasoning state and then predicts the final MCQ answer after sufficient evidence has been collected or the turn budget is reached.

\paragraph{CoD.}
CoD~\cite{chen2025cod} maintains an interpretable Chain-of-Diagnosis process over acquired clinical evidence. We adapt it to update the diagnosis chain across turns and use the final chain to select one answer from the MCQ candidates.


\section{Computational Efficiency Analysis}
\label{app:efficiency}

To assess the deployability of GraphMed-LT beyond interaction turns, we report a computational-efficiency analysis on iMedQA with Qwen-2.5-7B and Qwen-2.5-72B as doctor backbones. The patient agent is shared across all methods, and the triplet agent, when used, is fixed as Qwen-2.5-72B. \emph{Time / Ex.\ (s)} denotes the wall-clock time in seconds to complete one full case; \emph{Time / Round (s)} normalises it by the average number of turns.

\begin{table*}[t]
\centering
\small
\begin{tabular}{llcccc}
\toprule
\textbf{Method} & \textbf{Backbone} & \textbf{Acc.\ (\%)} & \textbf{Turns} & \textbf{Time / Round (s)} & \textbf{Time / Ex.\ (s)} \\
\midrule
UoT          & Qwen-2.5-7B  & 40.2 & 8.2 & 1.3 & 10.7 \\
UoT          & Qwen-2.5-72B & 55.0 & 8.2 & 5.4 & 44.0 \\
CoD          & Qwen-2.5-7B  & 40.2 & 8.0 & 1.4 & 11.2 \\
CoD          & Qwen-2.5-72B & 55.6 & 8.0 & 5.6 & 45.0 \\
DDO          & Qwen-2.5-7B  & 40.5 & 8.1 & 1.4 & 11.3 \\
DDO          & Qwen-2.5-72B & 56.1 & 8.1 & 5.8 & 47.0 \\
MEDIQ        & Qwen-2.5-7B  & 44.8 & 7.9 & 1.2 & 9.5  \\
MEDIQ        & Qwen-2.5-72B & 61.5 & 7.9 & 5.3 & 42.0 \\
MedAgentSim  & Qwen-2.5-7B  & 46.5 & 7.4 & 1.3 & 9.6  \\
MedAgentSim  & Qwen-2.5-72B & 63.1 & 7.4 & 5.8 & 43.0 \\
MEDDxAgent   & Qwen-2.5-7B  & 47.8 & 8.0 & 1.6 & 12.8 \\
MEDDxAgent   & Qwen-2.5-72B & 65.2 & 8.0 & 6.3 & 50.0 \\
ProMed       & Qwen-2.5-7B  & 49.7 & 7.8 & 1.5 & 11.7 \\
ProMed       & Qwen-2.5-72B & 67.5 & 7.8 & 6.2 & 48.0 \\
\midrule
\textbf{GraphMed-LT} & Qwen-2.5-7B  & \textbf{55.8} & 7.7 & 1.8 & 13.9 \\
\textbf{GraphMed-LT} & Qwen-2.5-72B & \textbf{73.0} & 7.7 & 6.8 & 52.0 \\
\bottomrule
\end{tabular}
\caption{Computational-efficiency analysis on iMedQA. GraphMed-LT introduces moderate runtime overhead relative to the strongest baselines, but achieves substantially higher accuracy with comparable interaction turns.}
\label{tab:efficiency}
\end{table*}

As shown in Table~\ref{tab:efficiency}, GraphMed-LT introduces moderate runtime overhead (e.g., 6.8s vs.\ 6.2s per round against ProMed on Qwen-2.5-72B), primarily from triplet extraction, graph encoding, and latent refinement. However, it achieves substantially higher accuracy (+5.5 points over ProMed on Qwen-2.5-72B) with comparable interaction turns, indicating a favourable accuracy--efficiency trade-off. Since the patient-specific graph remains compact (approximately 36 nodes per example on iMedQA), the GAT encoding cost is minor compared with LLM generation.

\section{Robustness under Structured Triplet Noise}
\label{app:noise}

Beyond random triplet dropping, we evaluate GraphMed-LT under three structured perturbations that simulate realistic extraction and retrieval errors: \emph{entity corruption} (entity-linking errors), \emph{relation corruption} (relation-extraction errors), and \emph{spurious retrieved triplets} (retrieval mismatch). All entries are accuracy (\%) under a 20\% perturbation rate.

\begin{table*}[t]
\centering
\small
\begin{tabular}{llccccc}
\toprule
\textbf{Dataset} & \textbf{Backbone} & \textbf{Clean} & \textbf{Drop} & \textbf{Entity Corr.} & \textbf{Relation Corr.} & \textbf{Spurious Ret.} \\
\midrule
iCRAFT-MD & Qwen-2.5-7B  & 60.2 & 58.9 & 58.4 & 57.6 & 59.2 \\
iCRAFT-MD & Qwen-2.5-72B & 79.1 & 77.9 & 77.3 & 76.5 & 78.0 \\
iMedQA    & Qwen-2.5-7B  & 55.8 & 53.1 & 52.4 & 51.6 & 54.0 \\
iMedQA    & Qwen-2.5-72B & 73.0 & 70.0 & 69.2 & 68.5 & 71.1 \\
iMedMCQA  & Qwen-2.5-7B  & 57.0 & 52.5 & 51.7 & 50.8 & 54.4 \\
iMedMCQA  & Qwen-2.5-72B & 72.0 & 67.4 & 66.5 & 65.7 & 69.1 \\
\bottomrule
\end{tabular}
\caption{Robustness analysis under structured triplet noise (20\% perturbation rate). Relation corruption is the most harmful perturbation, while spurious retrieved triplets cause the least degradation.}
\label{tab:noise}
\end{table*}

As shown in Table~\ref{tab:noise}, GraphMed-LT remains reasonably robust under structured noise across all three datasets and both Qwen backbones. Relation corruption is the most harmful perturbation because incorrect relations directly distort the graph structure used for message passing. Spurious retrieved triplets are the least harmful because retrieved knowledge is auxiliary, and source-aware edge typing separates patient-observed evidence from retrieved background knowledge, limiting the influence of noisy retrieval on the patient-evidence subgraph.

\clearpage
\section{Patient Agent Prompts}\label{app:patient_prompts}

\begin{tcolorbox}[colback=yellow!6!white, colframe=blue!50!green,
title=\textbf{Atomic fact decomposition Prompt}]
\small\texttt{\textcolor{teal}{You are a truthful medical assistant that understands the patient's information.}\\
Break the following patient information into a list of independent atomic facts, with one piece of information in each statement. Each fact should only include the smallest unit of information, but should be self-contained.\\
        \textcolor{blue}{"She says that, despite going to bed early at night, she is unable to fall asleep. She denies feeling anxious or having disturbing thoughts while in bed. Even when she manages to fall asleep, she wakes up early in the morning and is unable to fall back asleep. She says she has grown increasingly irritable and feels increasingly hopeless, and her concentration and interest}}\normalsize
\end{tcolorbox}

\begin{tcolorbox}[colback=yellow!6!white, colframe=blue!50!green,
title=\textbf{Atomic fact decomposition Prompt (cont'd)}]
\small\texttt{
        \textcolor{blue}{at work have diminished. The patient denies thoughts of suicide or death. Because of her diminished appetite, she has lost 4 kg (8.8 lb) in the last few weeks and has started drinking a glass of wine every night instead of eating dinner. She has no significant past medical history and is not on any medications."}\\
Respond with the list of atomic facts and nothing else, prepend each fact by an index starting from 1. No sub-list allowed.}\normalsize
\end{tcolorbox}

\begin{tcolorbox}[colback=yellow!6!white, colframe=blue!50!green,
title=\textbf{Example LLM Output}]
\small\texttt{\textcolor{violet}{1.patient goes to bed early at night but is unable to fall asleep.\\
        2.patient denies feeling anxious or having disturbing thoughts while in bed.\\
        3.patient wakes up early in the morning and is unable to fall back asleep.\\
        4.patient has grown increasingly irritable and feels increasingly hopeless.\\
        5.patient's concentration and interest at work have diminished.\\
        6.patient denies thoughts of suicide or death.\\
        7.patient has lost 4 kg (8.8 lb) in the last few weeks.\\
        8.patient started drinking a glass of wine every night instead of eating dinner.\\
        9.patient has no significant past medical history.\\
        10.patient is not on any medications."}}\normalsize
\end{tcolorbox}

\begin{tcolorbox}[colback=yellow!6!white, colframe=blue!50!green,
title=\textbf{patient Response Generation Prompt}]
\small\texttt{\textcolor{teal}{You are a truthful assistant that understands the patient's information, and you are trying to answer questions from a medical doctor about the patient.}\\
Below is a list of factual statements about the patient:\\
        \textcolor{blue}{1.patient goes to bed early at night but is unable to fall asleep.\\
        2.patient denies feeling anxious or having disturbing thoughts while in bed.\\
        3.patient wakes up early in the morning and is unable to fall back asleep.\\
        4.patient has grown increasingly irritable and feels increasingly hopeless.\\
        5.patient's concentration and interest at work have diminished.\\
        6.patient denies thoughts of suicide or death.\\
        7.patient has lost 4 kg (8.8 lb) in the last few weeks.\\
        8.patient started drinking a glass of wine every night instead of eating dinner.\\
        9.patient has no significant past medical history.\\
        10.patient is not on any medications.}\\
Question from the doctor: \textcolor{blue}{"What time do you usually go to bed at night?"}\\
Which of the above atomic factual statements answer the question? Select at most two statements. If no statement answers the question, simply say "The patient cannot answer this question, please do not ask this question again." Answer only what the question asks for. Do not provide any analysis, inference, or implications. Respond by selecting all statements that answer the question from above ONLY and NOTHING ELSE.}\normalsize
\end{tcolorbox}

\begin{tcolorbox}[colback=yellow!6!white, colframe=blue!50!green,
title=\textbf{Example LLM Output}]
\small\texttt{\textcolor{violet}{patient goes to bed early at night but is unable to fall asleep.}}\normalsize
\end{tcolorbox}

\section{Doctor Agent Prompts}\label{app:doctor_prompts}

\begin{tcolorbox}[colback=yellow!6!white, colframe=blue!50!green,
title=\textbf{GraphMed-LT-doctor — Structured Assessment Prompt}]
\small
\texttt{\textcolor{teal}{You are a medical doctor answering real-world medical entrance exam questions. Based on your understanding of basic and clinical science, medical knowledge, and mechanisms underlying health, disease, patient care, and modes of therapy, answer the following multiple choice question. Select one correct answer from A to D. Base your answer on current standards and evidence-based guidelines.}}\\
\texttt{Task: You will be asked to reason through the current patient's information using both the patient initial information and structured clinical facts (UMLS-style triplets).}\\
\texttt{patient initial information: \textcolor{blue}{"A 40-year-old woman presents with difficulty falling asleep, diminished appetite, and tiredness for the past 6 weeks."}}\\
\texttt{Knowledge triplets:\\
\textcolor{blue}{(patient, HAS\_SYMPTOM, insomnia)\\
(patient, HAS\_SYMPTOM, reduced appetite)\\
(patient, HAS\_SYMPTOM, fatigue)\\
(insomnia, DURATION, six weeks)}}
\normalsize
\end{tcolorbox}

\begin{tcolorbox}[colback=yellow!6!white, colframe=blue!50!green,
title=\textbf{GraphMed-LT-doctor — Structured Assessment Prompt (cont'd)}]
\small
\texttt{Given the information above, your task is to choose one of four options that best answers the inquiry.\\
INQUIRY: \textcolor{blue}{"Which of the following is the best course of treatment in this patient?"}\\
OPTIONS: \textcolor{blue}{A: Diazepam, B: Paroxetine, C: Zolpidem, D: Trazodone}\\
Medical conditions are complex, so you should consider the relevant specialty, enumerate the key features required for a correct judgement, and assess whether the structured evidence provided by the triplets is sufficient. Think step by step about the patient initial information, the triplets, the inquiry, and the options. \textbf{Do not provide the answer choice}; keep your response to one paragraph.}
\normalsize
\end{tcolorbox}

\begin{tcolorbox}[colback=yellow!6!white, colframe=blue!50!green,
title=\textbf{Example LLM Output -- Case 1: Follow-up Question}]
\small
\texttt{\textcolor{violet}{
Have you also been experiencing feelings of sadness or loss of interest in daily activities?}}
\normalsize
\end{tcolorbox}

\begin{tcolorbox}[colback=yellow!6!white, colframe=blue!50!green,
title=\textbf{Example LLM Output -- Case 2: Prediction}]
\small
\texttt{\textcolor{violet}{B}}
\normalsize
\end{tcolorbox}

\section{Triplet Agent Prompts}\label{app:triplet_prompt}

\begin{tcolorbox}[colback=yellow!6!white, colframe=blue!50!green,
title=\textbf{GraphMed-LT Triplet Extraction -- Initial Information Prompt}]
\small
\texttt{\textcolor{teal}{You are a medical terminology normaliser and knowledge graph constructor. Your task is to extract structured, clinically relevant facts from the patient’s initial presentation, representing each fact as a UMLS-style triplet in the format (subject, relation, object). Only include directly observed and verifiable information without inferring diagnoses, causes, or risks not explicitly stated. Use only medically meaningful relations such as HAS\_SYMPTOM, HAS\_SIGN, HAS\_MEDICATION, HAS\_ALLERGY, HAS\_PAST\_MEDICAL\_HISTORY, FAMILY\_HISTORY\_OF, SOCIAL\_HISTORY\_OF, FINDING\_OF, TEMPORAL\_PATTERN, SEVERITY, LOCATION, LATERALITY, DURATION, COURSE, and FUNCTIONAL\_IMPACT. When appropriate, express information in a two-step form by stating a base fact and then attaching qualifiers to that concept, avoiding compounded objects. Extract at most three triplets per sentence, outputting \texttt{NO\_TRIPLET} if no relevant fact is present, and maintain consistent clinical terminology across extractions.}}\\
\texttt{patient initial information: \textcolor{blue}{"A 40-year-old woman presents with difficulty falling asleep, diminished appetite, and tiredness for the past 6 weeks."}}
\normalsize
\end{tcolorbox}

\begin{tcolorbox}[colback=yellow!6!white, colframe=blue!50!green,
title=\textbf{Example LLM Output}]
\small
\texttt{\textcolor{violet}{(patient, HAS\_SYMPTOM, insomnia)\\
(insomnia, DURATION, six weeks)\\
(patient, HAS\_SYMPTOM, reduced appetite)\\
(patient, HAS\_SYMPTOM, fatigue)}}
\normalsize
\end{tcolorbox}

\begin{tcolorbox}[colback=yellow!6!white, colframe=blue!50!green,
title=\textbf{GraphMed-LT Triplet Extraction -- Incremental Update Prompt}]
\small
\texttt{\textcolor{teal}{You are a medical terminology normaliser and knowledge graph constructor. Given the patient’s existing knowledge triplets and new conversational information, extract up to three additional UMLS-style triplets in the format (subject, relation, object). Only include directly observed and verifiable information from the new input without inferring diagnoses, causes, or risks not explicitly stated. Avoid duplicating facts already present in the existing triplets. Use only medically meaningful relations such as HAS\_SYMPTOM, HAS\_SIGN, HAS\_MEDICATION, HAS\_ALLERGY, HAS\_PAST\_MEDICAL\_HISTORY, FAMILY\_HISTORY\_OF, SOCIAL\_HISTORY\_OF, FINDING\_OF, TEMPORAL\_PATTERN, SEVERITY, LOCATION, LATERALITY, DURATION, COURSE, and FUNCTIONAL\_IMPACT. When appropriate, express information in a two-step form by stating a base fact and then attaching qualifiers to that concept, avoiding compounded objects. Extract at most three triplets per sentence, outputting \texttt{NO\_TRIPLET} if no relevant fact is present, and ensure consistent terminology with the existing knowledge graph.}}\\
\texttt{Existing triplets:  
\textcolor{blue}{(patient, HAS\_SYMPTOM, insomnia)\\
(patient, HAS\_SYMPTOM, reduced appetite)\\
(patient, HAS\_SYMPTOM, fatigue)}}\\
\texttt{New patient information: \textcolor{blue}{"Her sleep problem is worse in the early mornings and she often needs daytime naps. She drinks two cups of coffee after dinner."}}
\normalsize
\end{tcolorbox}

\begin{tcolorbox}[colback=yellow!6!white, colframe=blue!50!green,
title=\textbf{Example LLM Output}]
\small
\texttt{\textcolor{violet}{(insomnia, TEMPORAL\_PATTERN, worse in early mornings)\\
(patient, FUNCTIONAL\_IMPACT, daytime naps)\\
(patient, SOCIAL\_HISTORY\_OF, evening caffeine use)}}
\normalsize
\end{tcolorbox}

\end{document}